# Monitoring Indoor Activity of Daily Living Using Thermal Imaging: A Case Study

Hassan M. Ahmed[1], Bessam Abdulrazak[1]
[1] AMI-Lab
Faculté des sciences, Université de Sherbrooke

*Abstract*—Monitoring indoor activities of daily living (ADLs) of a person is neither an easy nor an accurate process. It is subjected to dependency on sensor type, power supply stability, and connectivity stability without mentioning artifacts introduced by the person himself. Multiple challenges have to be overcome in this field, such as; monitoring the precise spatial location of the person, and estimating vital signs like an individual's average temperature. Privacy is another domain of the problem to be thought of with care. Identifying the person's posture without a camera is another challenge. Posture identification assists in the person's fall detection. Thermal imaging could be a proper solution for most of the mentioned challenges. It provides monitoring both the person's average temperature and spatial location while maintaining privacy. In this research, we propose an IoT system for monitoring an indoor ADL using thermal sensor array (TSA). Three classes of ADLs are introduced, which are daily activity, sleeping activity and no-activity respectively. Estimating person average temperature using TSAs is introduced as well in this paper. Results have shown that the three activity classes can be identified as well as the person's average temperature during day and night. The person's spatial location can be determined while his/her privacy is maintained as well.

## I. INTRODUCTION

Monitoring indoor activities of daily living (ADLs) can be achieved with different methods [1]–[5]. Utilizing motion sensors is one of them [6]–[8]. Motion sensors can be used to detect if there is any motion taking place in front of the sensor. On the one hand, these sensors are effective for detecting the movement of one person, and their effectiveness increases when sensor fusion is deployed. However, they impose several limitations that can be summarized as follows: a) they cannot detect the exact spatial location of the monitored person or even an estimate for its spatial location, b) they cannot differentiate between the steady sitting state and the no-motion state, c) they cannot be used for fall detection as they cannot obtain depth and 2D data about the person's location, and d) they do not inform about the number of individuals inside the room. Moreover, this type of sensor has another limitation is that it does not give any vital data about the person during the monitoring period.

Another option of detecting ADLs is using another physical quantity i.e., the temperature of the person [9]–[12]. Every human being can be considered as a heat source that can be detected using Thermal Sensor Array (TSA).

Thus, the problem can be reduced to be only the detection of the heat distribution of the person inside the room. In other words, we can track the estimated spatial location of the person inside the room by tracking his/her thermal distribution. (Fig. 1 shows the thermal 2D distribution of two heat sources as measured by a thermal sensor array).

Given the various advantages of thermal imaging, the contribution of this paper is to assess the effectiveness of a thermal sensor array for tracking a subject's indoor activities.

The paper is organized as follows: section II discusses the advantages of thermal imaging over the mentioned limitations. Section III methodology used in conducting the experiment. Results of the experiment and related discussion are introduced in section IV. Finally, the conclusion is presented in section V.

## II. THERMAL IMAGING TECHNOLOGY

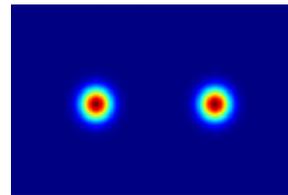

Fig. 1. Thermal distribution of two heat sources as picked up by an array sensor after interpolation.

A thermal sensor array can be used to pick up a complete capture for the thermal distribution of the room every minute and store its readings for subsequent processing [13], [14] [15]. This method has many advantages. First, it maintains the privacy of participating persons. Second, it tracks the precise spatial location of the person, giving rise to a better understanding of behavior patterns extracted from activity per spatial location. Third, it monitors the average temperature of the person per unit timestamp allowing for better assessment of the person's health status (e.g., if he/she is suffering from fever or relevant diseases). Fourth, it can track whether the person is in an upstanding posture or has lain down, using advanced algorithms that can differentiate between the thermal distribution in both cases. Upstanding posture gives more concise (confined) thermal distribution for the person, alike the horizontal or lying down position where relevant thermal distribution is much wider and scattered. This feature can also be helpful in fall detection [16].

Several researches have concluded the possibility of indoor occupancy estimation using TSA [13], [14]. They all placed TSA in a specific spatial location in the room. The person IR emission is triangulated to estimate his location as in [15]. Alike their work which focused on occupancy detection and estimation only without giving attention to the person's ADL indoors, in our work we focus on monitoring a person's ADL to deduce his behavior. Following, we discuss the methodology used in our IoT experiment.

### III. METHODOLOGY

A male person is monitored over time using a thermal sensor array to track his activity. We are monitoring a single person as a proof of concept and for feasibility. We implemented an IoT system consisting of a thermal sensor array and a processing unit to analyze the acquired data. The person's activity is monitored by tracking specific thermal image pixels related to the activity spatial locations inside the room. We track the temperature of the corresponding pixels in the temporal domain to construct different activity vectors/arrays for the person. Then these vectors are sent to a cloud server annotated with their relevant timestamp, where they can be stored and analyzed.

Our system is deployed to monitor a single person living inside a room. The person's activity is monitored on a 24-hour basis and is classified as three classes: the sleeping activity, the daily activity and the no-activity classes. The corresponding spatial locations at which these three activities are most likely to happen are marked on the room schematic. The bed represents the sleeping activity of the person. The working table and dining table represents the daily activity of the subject. The room schematic is presented in Fig. 3 along with the sensor located inside the room and its location with respect to the person. The activities' spatial locations are presented in Fig. 4.

The IoT system consists of MLX90641 (16x12/110°x75°/~7m axial range) [17] which is a thermal sensor array for acquiring thermal activity and a Raspberry Pi 3B+ is used as a node. The acquired thermal frame is transmitted as a 1D-array (element-by-element) along with a corresponding timestamp via CoAP transmission protocol from the Pi to a server for subsequent data processing and analysis. Another copy from the transmitted data is stored locally on the Pi for retrieval and substitution in case of transmission failure situations. The IoT system block diagram is presented in Fig. 5. The complete process of the system is presented in Fig. 6. The experimental hardware setup is shown in Fig. 7. The next section explains the temporal thermal activity tracking algorithm.

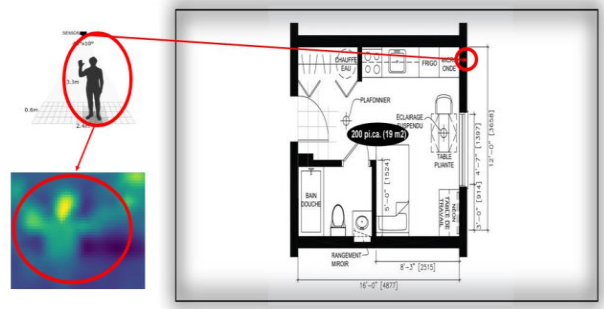

Fig. 3. Room schematic along with its real photo, describing the spatial location of the sensor w.r.t the person body.

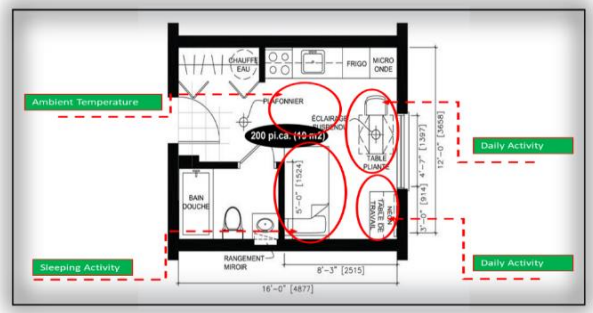

Fig. 4. Regions of interest along with its corresponding activity.

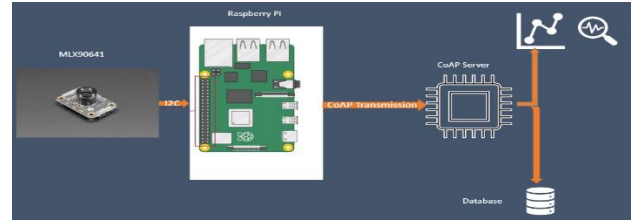

Fig. 5. Block diagram of the complete system.

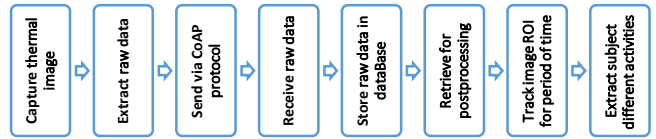

Fig. 6. Complete System Process.

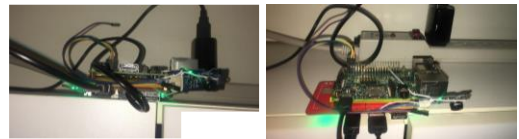

Fig. 7. Complete hardware setup during runtime.

#### A. The Temporal Thermal Activity Tracking Algorithm

This algorithm is used to track the spatial thermal activity in the acquired image from the sensor array. It starts with tagging the corresponding pixels that are relevant to the spatial locations where the three activity classes are most likely to happen. The tagging is performed by relating specific image pixels to a location where each class of the three classes happen, i.e. for the sleeping activity class, the person is asked to lie down on the bed and a reference image is captured to identify the location of the bed and so on for the other two classes. Then the related pixels to these locations are tracked over time. This process is shown in Fig. 8.

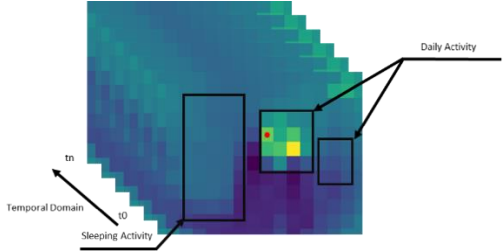

Fig.8 Visualization of temporal thermal activity tracking algorithm.

Each thermal activity corresponding to one of the three classes can be formulated mathematically as a 1D-array by the following formula (1):

$$A^K = [a_t^K, a_t^K, a_t^K, \ldots\ldots, a_T^K], \forall\, t = \{0,1,2,\ldots,T\} \text{ and } K \in \{1,2,3\} \quad (1)$$

Where *A* is the thermal activity array, *t* is the temporal resolution, *T* is the final timestamp of the monitoring period, and *K* is the activity class (1: Sleeping, 2: Daily and 3: No-activity).

## IV. RESULTS AND DISCUSSION

The non-interpolated image (16x12) as captured from the sensor, its corresponding interpolated image (128x176) and no-activity image are presented in Fig. 9 (a), (b) and (c) respectively. The person's location inside the image is labeled by the red dot. Two different ways of sleeping activities are presented in Fig. 10.

We have monitored the person's behavior for 10 consecutive days in April 2021. Every monitored day is presented in a separate curve starting from 00:00 o'clock till 11:59 of the same day on the X-axis and the temperature on the Y-axis, except for the last day where the monitoring ended at 16:30. The orange curve represents sleeping activity and the blue curve represents daily activity for the monitored person. The first three days of monitoring are shown in Fig. 11, namely 7[th], 8[th] and 9[th] April. The three consecutive days which are 10[th], 11[th] and 12[th] April are shown in Fig. 12. Fig. 13 shows the person's behavior during 13[th], 14[th] and 15[th] April. Finally, the last two days are shown in Fig. 14.

Sleeping activity periods, daily activity periods, no-activity periods, and missing data periods for each day are listed in Table 1 and are shown as a bar chart in Fig. 15. Behavior statistics such as mean value and percentage value for each activity type along the whole monitoring period are listed in Table 2.

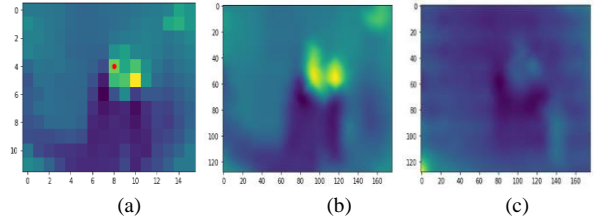

(a) (b) (c)

Fig. 9. (a) non-interpolated image showing the person doing his daily activity, (b) the corresponding interpolated image, and (c) interpolated image showing no activity inside the room.

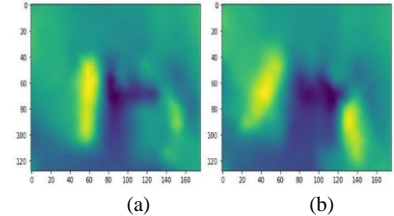

(a) (b)

Fig. 10. Two different sleeping activities of the person.

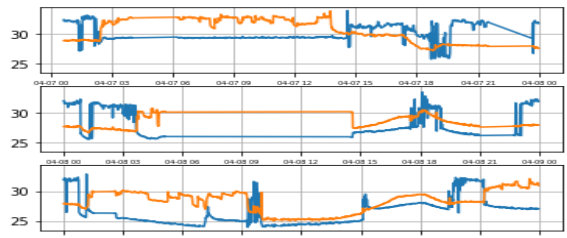

Fig. 11. Monitoring person's behavior during 7[th], 8[th] and 9[th] April.

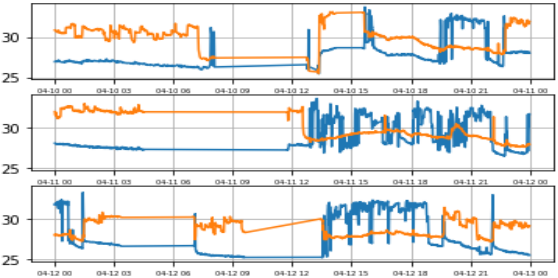

Fig. 12. Monitoring person's behavior during 10[th], 11[th] and 12[th] April.

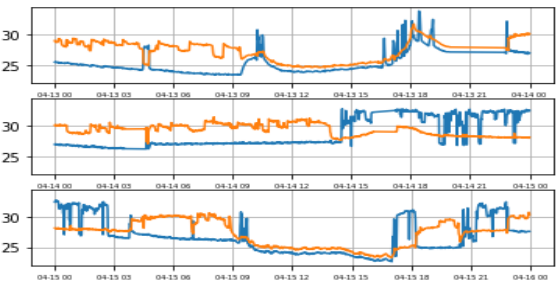

Fig. 13. Monitoring person's behavior during 13[th], 14[th] and 15[th] April.

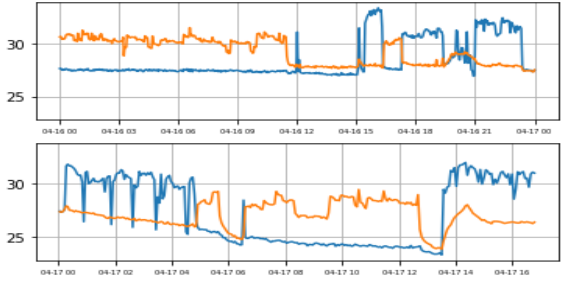

Fig. 14. Monitoring person's behavior during 16th and 17th April.

TABLE 1. BEHAVIOR MONITORING PERIODS.

| Activity Period (Hours) | Monitoring Day | | | | | | | | | | |
|---|---|---|---|---|---|---|---|---|---|---|---|
| | 7th | 8th | 9th | 10th | 11th | 12th | 13th | 14th | 15th | 16th | 17th |
| Daily Activity | 11.5 | 6.5 | 4 | 5 | 6 | 6 | 4 | 10.5 | 7 | 8 | 9.5 |
| Sleeping Activity | 11.5 | 2 | 10 | 12 | 5 | 7.5 | 10 | 13 | 9.5 | 11.5 | 8 |
| No Activity | 1 | 6.5 | 10 | 2 | 5 | 2 | 7 | 0.5 | 7.5 | 4.5 | 1 |
| Missing Data | 0 | 9 | 0 | 5 | 8 | 8.5 | 3 | 0 | 0 | 0 | 5.5 |

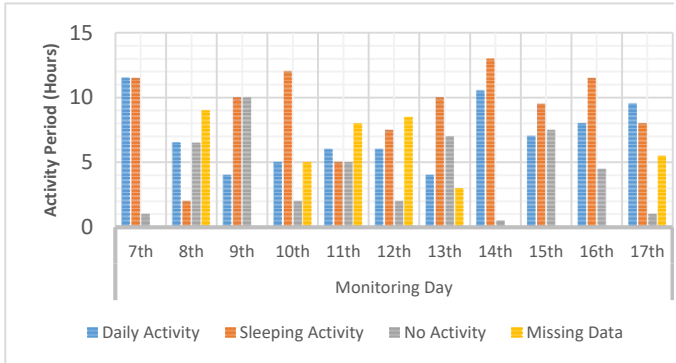

Fig. 15. Different activity periods for the person's behavior during the whole monitoring period.

TABLE 2. BEHAVIOR STATISTICS FOR THE WHOLE MONITORING PERIOD.

| Activity Type | Mean (Hours) | Percentage |
|---|---|---|
| Daily Activity | 7.090909 | 0.295454545 |
| Sleeping Activity | 9.090909 | 0.378787879 |
| No Activity | 4.272727 | 0.178030303 |
| Missing Data | 3.545455 | 0.147727273 |

On each day of the monitoring period, the sleeping activity and the daily activity are observed to be contrary to each other. Both activities have a temperature difference of about 3°C degrees each day. The person's average temperature is observed to be 32°C degrees on average during the whole monitoring period.

The most dominant activity of the person during the monitoring period is the sleeping activity as concluded from the behavior statistics in Table 2 with an average of 9 hours per day. Daily activity happens to be the second place with 7 hours per day, and the no activity comes in third place with about 4.2 hours per day.

Small periods of no-activity happening inside the room are inferred as a bathroom entry due to its small duration and its location between two long duration of sleeping activity and daily activity. The person is most likely considered visiting the bathroom just after wake up. The person's concluded bathroom visits are listed in Table 3.

TABLE 3. THE PERSON'S BATHROOM VISITS.

| Day | 7th | 8th | 9th | 10th | 11th | 12th | 13th | 14th | 15th | 17th |
|---|---|---|---|---|---|---|---|---|---|---|
| Bathroom Visit (start / duration [minutes]) | 14:00 / 60 | 01:00 / 30 | 01:30 / 30 | 07:00 / 30 | 17:30 / 30 | 01:30 / 30 | 09:30 / 30 | 14:15 / 30 | 03:00 / 30 | 12:30 / 60 |
| | | | | 13:00 / 30 | 22:00 / 30 | 18:30 / 30 | | | | |
| | | | | 21:30 / 60 | 22:30 / 90 | 21:00 / 60 | | | | |

We conclude from Table 3 that the person's bathroom visit takes on average between 30 minutes to 60 minutes.

During the long no-activity durations exceeding one hour the person is considered out of the room, i.e., these durations are considered outing periods. On 9th, 13th, 15th and 16th we observe long durations of no-activity at the same normal working hours, hence we conclude that the person was out for work at these periods.

We can also notice that the person often goes to sleep after midnight and before 03:00 except for 8th, 15th and 17th April, where the person went to bed after 03:00. We conclude here that the person is a night owl.

We can notice that the person spends about 29% of its day as daily activity, 37% as sleeping activity, and 17% as no-activity or bathroom visits and outings. Missing data is about 14%.

In terms of monitoring the person's temperature, a significant decrease in the person's body temperature is observed during 12th April around 20:00 and during 17th April from 00:00 till around 05:00.

V. CONCLUSION

We proposed and implemented in this paper a monitoring system based on thermal sensor array that can capture a person's activities of daily living (ADLs). The monitored ADLs are classified as sleeping, daily, and no-activity at all. The experiment proves that the system enables detection of a person's spatial location indoor precisely. In addition, the experiment enables prediction for the bathroom visits and the outing and estimates the person's temperature during the whole monitoring period along with maintaining the person's privacy as well.